\documentclass[runningheads]{llncs}
\usepackage{amssymb}
\usepackage{amsmath}
\usepackage{graphicx}
\usepackage{hyperref}
\usepackage[caption=false]{subfig}
\usepackage{xcolor}
\graphicspath{{figures/}}

\begin{document}
\title{Learning Representations of Endoscopic Videos to Detect Tool Presence Without Supervision}
\titlerunning{Representations of Endoscopic Videos}        

\author{David Z. Li \inst{1}\and
Masaru Ishii \inst{2} \and
Russell H. Taylor \inst{1} \and
Gregory D. Hager \inst{1}\and
Ayushi Sinha \inst{3}}

\authorrunning{Li et al.} 

\institute{Department of Computer Science, The Johns Hopkins University, USA\\
            \email{dli44@alumni.jhu.edu}
            \and
            Johns Hopkins Medical Institutions, USA
            \and 
            Laboratory for Computational Sensing and Robotics, The Johns Hopkins University, USA
}

\maketitle              
\begin{abstract}
In this work, we explore whether it is possible to learn representations of endoscopic video frames to perform tasks such as identifying surgical tool presence without supervision. We use a maximum mean discrepancy (MMD) variational autoencoder (VAE) to learn low-dimensional latent representations of endoscopic videos and manipulate these representations to distinguish frames containing tools from those without tools. 
We use three different methods to manipulate these latent representations in order to predict tool presence in each frame.
Our fully unsupervised methods can identify whether endoscopic video frames contain tools with average precision of 71.56, 73.93, and 76.18, respectively, comparable to supervised methods.
Our code is available at \url{https://github.com/zdavidli/tool-presence/}.

\end{abstract}
\section{Introduction}
\begin{figure}[t]
    \centering
    \includegraphics[width=\textwidth]{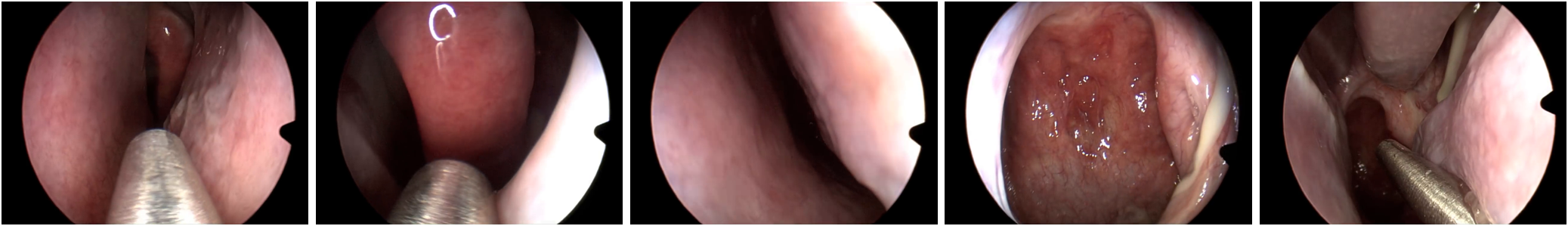}
    \caption{Examples of endoscopic video frames with and without tools. These examples show the variance in anatomy in our dataset.}
    \label{fig:examples}
\end{figure}

Despite the abundance of medical image data, progress in learning from such data has been impeded by the lack of labels and the difficulty in acquiring accurate labels. With increase in minimally invasive procedures~\cite{Tsui13}, an increasing number of endoscopic videos (Fig.~\ref{fig:examples}) are available. This can open up the opportunity for video-based surgical education and skill assessment. Prior work~\cite{Malpani15} has shown that both experts and non-experts can produce valid objective skill assessment via pairwise comparisons of surgical videos. However, watching individual videos is time consuming and tedious. Therefore, much work is being done in automating skill assessment using supervised~\cite{Dipietro16} and unsupervised~\cite{Dipietro18} learning. These prior methods used kinematic data from tools to learn surgical motion. However, many endoscopic procedures do not capture kinematic data. Therefore, we want to explore whether we can work towards automated skill assessment directly from endoscopic video.

Since videos contain more information than just kinematics, we want to first isolate tool motion from camera motion in endoscopic videos. If the two types of motion can be disentangled, then representations of video frames with and without tools should be distinct enough to allow separation between the two. Therefore, our aim in this work is to evaluate whether we can detect frames that contain tools. In order to do this, we use a variational autoencoder (VAE)~\cite{kingma2013auto} to learn latent representations of endoscopic video frames since VAEs have the ability to learn underlying low-dimensional latent representations from complex, high-dimensional input data.
Specifically, we use maximum mean discrepancy (MMD) VAE~\cite{Zhao2017InfoVAEIM}, which uses a modified objective function in order to avoid overfitting and promote learning a more informative latent representation.

We then manipulate these learned representations or encodings using three different methods. First, we directly use the encodings produced by our MMD-VAE to evaluate whether encodings of frames with and without tools can be separated. Second, we model the tool presence as a binary latent factor and train a Bayesian mixture model to learn the clusters over our encodings and classify each frame as containing or not containing tools. Third, we use sequences of our encodings to perform future prediction and evaluate whether temporal context can better inform our prediction of tool presence. 
Our evaluation methods identify frames containing tools with average precision of $71.56$, $73.93$, and $76.18$, respectively, without any explicit labels.

\section{Prior work}

Prior work has shown that surgical motion can be learned from robot kinematics. 
Lea et al.~\cite{Lea16} and DiPietro et al.~\cite{Dipietro16} showed that supervised learning methods can accurately model and recognize surgical activities from robot kinematics. 
DiPietro et al.~\cite{Dipietro18} further showed that encodings learned from robot kinematics in an unsupervised manner also clustered according to high-level activities. 
However, these methods rely on robot kinematics which provide information like gripper angle, velocity, etc. Endoscopic procedures that do not use robotic manipulation do not produce kinematics data, but do produce endoscopic videos.

Much work has also been dedicated to extracting tools from video frames
using supervised tool segmentation and tool tracking methods. 
Many methods ignore the temporal aspect of videos and compute segmentation on a frame-by-frame basis. 
Several methods use established network architectures, like U-Net, to compute segmentations~\cite{Pakhomov17,Shvets18}. 
Some methods have tried to tie in the temporal aspect of videos by using recurrent neural networks (RNNs) to segment tools~\cite{Attia17}, while others have combined simpler fully convolutional networks with optical flow to capture the temporal dimension~\cite{Herrera17}. 
More recently, unsupervised methods for learning representations of videos have also been presented~\cite{Srivastava15}. While Srivastava et al.~\cite{Srivastava15} use RNNs to encode sequences of video frames, which can grow large quickly, we will explore whether unsupervised learning of video representations on a per-frame basis will give us sufficient information to discriminate between frames with and without tools. 

\section{Method}

\subsection{Dataset}
We use a publicly-available sinus endoscopy video consisting of five segments of continuous endoscope movement from the front of the nasal cavity to the back~\cite{ephrat}. The video was initially collected at 1080p resolution and split into individual frames.
Frames that depicted text or where the endoscope was outside the nose were discarded. A total of $1551$ frames, downsampled from 1080p resolution to a height of 64 pixels and centrally cropped to $64\times 64$ pixels, were extracted. 
This downsampling was necessary due to GPU limitations. 

We held out $20\%$ of the frames, sampled from throughout our video sequence, as the test set. 
Each frame was manually labeled for tool presence.
These annotations were used for evaluation only. 
In the training set, $65.8\%$ of frames were labeled as containing a tool,
and in the test set $67.4\%$ of frames were labeled as containing a tool.

\subsection{Variational autoencoder}

We use variational autoencoders (VAEs)~\cite{kingma2013auto} to learn low dimensional latent representations that can encode our endoscopic video data. VAEs and their extensions~\cite{Zhao2017InfoVAEIM} are based on the idea that each data point, ${x} \in \mathbf{X}$, is generated from a $d$-dimensional latent random variable, $z\in \mathbb{R}^d$
, with probability $p_{\theta}\,({x} | {z} )$, where $z$ is sampled from the prior, $p_{\theta}(z) \sim \mathcal{N}(\mu,\,\sigma^{2})$, parameterized by $\theta$~\cite{kingma2013auto}. However, since optimizing over the probability density function (PDF) $P$ is intractable,
the optimization is solved over a simpler PDF, $Q$, to find $q_{\phi}$ that best approximates $p_{\theta}$~\cite{kingma2013auto}. To ensure that $q$ best approximates $p$, $\theta$ and $\phi$ are jointly optimized by maximizing the evidence lower bound (ELBO)~\cite{kingma2013auto}:
\begin{equation}
  \log p(x) \geq {E}_{q_{\phi}(z | x)}\left[\log p_{\theta}(x | z)\right]-\mathrm{KL}\left(q_{\phi}(z | x)\, \| \,p(z)\right).
\label{eq:vae_objective}
\end{equation}

In order to encourage VAEs to learn more informative encodings without overfitting to the data, Zhao et al.~\cite{Zhao2017InfoVAEIM} introduced the maximum mean discrepancy (MMD) VAE, which maximizes the mutual information between the data and the encodings. 
MMD-VAE changes the objective function by replacing the KL-divergence term with MMD and introduces a regularization term, $\lambda$~\cite{Zhao2017InfoVAEIM}:
\begin{equation}
  \log p(x) \geq {E}_{q_{\phi}(z | x)}\left[\log p_{\theta}(x | z)\right]-\lambda\mathrm{MMD}\left(q_{\phi}(z | x)\, \| \,p(z)\right).
\label{eq:infovae_objective}
\end{equation}

\subsection{Training}

\begin{figure}[t]
    \centering
    \includegraphics[width=\textwidth]{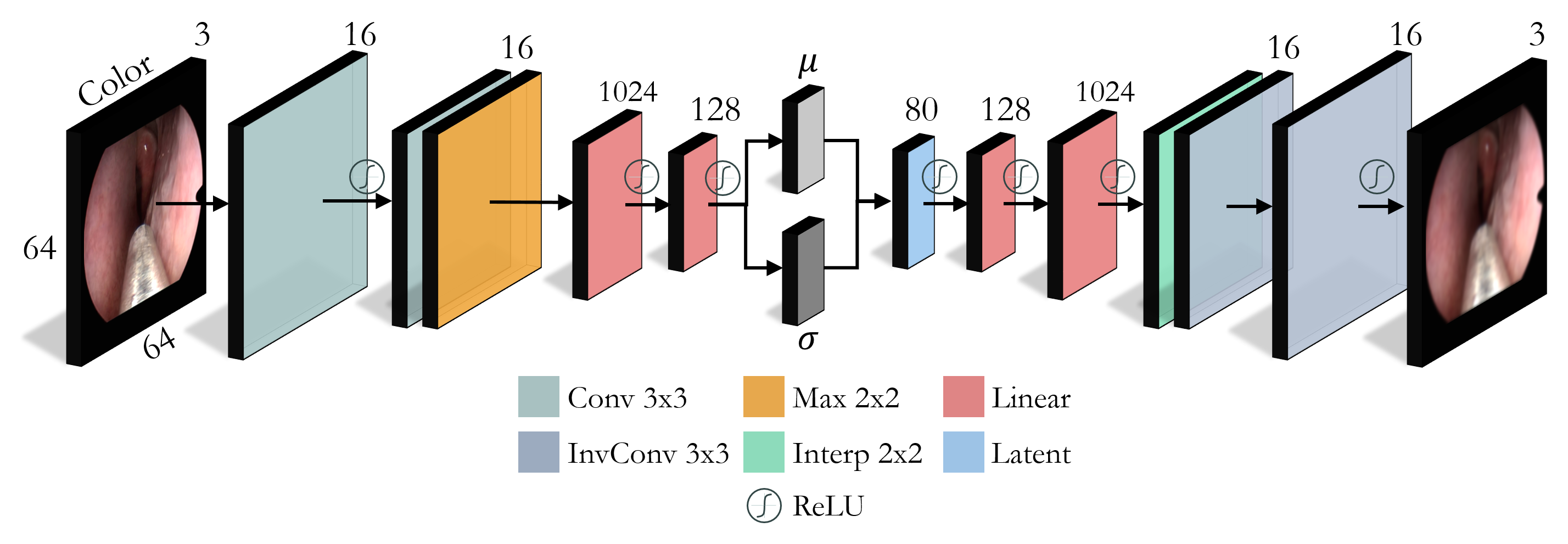}
    \caption{Our MMD-VAE architecture with a two-layer CNN encoder and decoder.}
    \label{fig:architecture}
\end{figure}

We used a convolutional neural network encoder and decoder each with two convolutional
layers and three fully-connected layers (Fig.~\ref{fig:architecture}).
We performed a hyperparameter sweep over latent dimension and regularization coefficient of our MMD-VAE implementation in PyTorch~\cite{paszke2017automatic} and evaluated each model based on the criteria in Section \ref{evaluation}.

The best performing model from our sweep was trained for 80 epochs using stochastic gradient descent (SGD) on a single
NVIDIA Quadro K620 GPU with 2GB memory with minibatch size of 32 and Adam optimizer~\cite{kingma2014adam} with default parameters except for learning rate which was set to $10^{-3}$. This model had a latent dimension $d=20$ and regularization coefficient $\lambda=5$.

\subsection{Model Evaluation}
\label{evaluation}

\subsubsection{Direct Evaluation.} First, we directly evaluate the encodings produced by our MMD-VAE implementation using a query-based evaluation. We compute the cosines between encodings of each test frame, $i$, containing a tool and all other test frames, $j \neq i$, and threshold the products to separate high and low responses.
Since the query (i.e., test frame, $i$) contains a tool, all other test frames containing tools should produce high response, while those without tools should produce low response. 
To evaluate our results, we compute the average precision (AP) score~\cite{zhu2004recall} over responses from each test frame. AP summarizes the precision-recall curve by computing the weighted mean of precision values computed at each threshold, weighted by the increase in recall from the previous threshold. In simpler terms, AP computes the area under the precision-recall curve.

\subsubsection{Approximate Inference.} Next, we evaluate our encodings using approximate inference. We estimate $p\,(\text{tool presence}\,|\,\text{latent encoding})$ by modeling the space of encodings as a finite mixture model with a categorical latent variable $C$ which has $K=2$ Gaussian states (tool present and not present). We use Markov chain Monte Carlo (MCMC) sampling~\cite{murphy2012machine} to approximate the posterior distribution by constructing a Markov chain whose states are assignments of the model parameters and whose stationary distribution is $p$: 
\begin{equation}
\label{eq:approxinf}
  p(z_i | \boldsymbol{\theta}, \mu_{c_i}, \sigma_{c_i}) = \sum_{c_i=1}^K \boldsymbol{\theta}_{c_i}^{\top} \mathcal{N}(z_i | \mu_{c_i}, \sigma_{c_i}).
\end{equation}
Here, $z_i\in \mathbb{R}^d$ is the $i$th encoding generated from our MMD-VAE, $d\in \mathbb{N}$ is the dimension of the encoding sample, and each of the $K$ configurations are normally distributed according to parameters $\boldsymbol{\mu}, \boldsymbol{\sigma}\in\mathbb{R}^K$ and mixing probabilities $\boldsymbol{\theta}\in \mathbb{R}^{d\times K}$. 
We assume each encoding $z_i$ is generated by $\boldsymbol{\theta}$ and a latent state $1\le c_i \le K$, described by $\mathcal{N}(\mu_{c_i}, \sigma_{c_i})$.
By running the Markov chain for $B$ burn-in steps, we reach the stationary distribution, $p$. 

We then sample the chain for $N$ iterations which form samples from $p$~\cite{murphy2012machine}.
The parameters of the mixture model, $\boldsymbol{\mu}, \boldsymbol{\sigma}, \text{ and } \boldsymbol{\theta}$, are learned using the No-U-turn sampler~\cite{hoffman2014no} on Eq.~\ref{eq:approxinf}.
Finally, for a fixed $z_i$, we can estimate the probability that encoding $z_i$
comes from latent cluster $c_i$ to predict tool presence: 
$p(c_i | z_i) \propto p(z_i|c_i)p(c_i) = p(z_i, c_i)$.

For evaluation, we learned the parameters in PyStan~\cite{pystan}
with four Markov chains with $B=N=2500$ and default hyperparameters. 
The posterior probability of each sample belonging to each cluster was then computed and used to predict labels. 
We evaluate the separation between the two latent states and compute the AP score for our label predictions.

\subsubsection{Future Prediction.} Finally, we evaluate our encodings by training a future prediction model using a recurrent neural network (RNN) encoder-decoder~\cite{cho14} to observe a sequence of past video frame encodings and reconstruct a sequence of future frame encodings~\cite{Dipietro18}. The intuition behind this approach is that models capable of future prediction must encode contextually relevant information~\cite{Dipietro18}. Both the encoder and decoder have long short-term memory (LSTM)~\cite{Gers00,Hochreiter97} architectures to avoid the vanishing gradient problem, and each frame of the future sequence is associated with its own mixture of multivariate Gaussians in order not to blur distinct futures together under a unimodal Gaussian~\cite{Dipietro18}.

Our PyTorch~\cite{paszke2017automatic} implementation of the future prediction model was similar to that presented by DiPietro et al.~\cite{Dipietro18}. We used 5 frame sequences of past and future encodings, and Adam~\cite{kingma2014adam} for optimization at a learning rate of $0.005$ and other hyperparameters at their default values. The latent dimension was set to $64$, the number of Gaussian mixture components to $16$, and the model was trained for $1000$ epochs with a batch size of $50$.

As in direct evaluation, we evaluate the encodings produced by future prediction by computing the cosines between encodings from each test sequence, $s_i$, containing a tool and all other test sequences, $s_j \neq s_i$, taking the maximum per-frame, and thresholding, as before, to separate high and low responses. A per-frame maximum is computed here since each frame belongs to multiple adjacent sequences, $s_j$, producing multiple responses. Specifically, since we used 5 frame sequences, each frame produces 5 responses, of which we pick the maximum.
Finally, we compute the AP over responses from each test sequence.

\section{Results}

The results from our three experiments are described in this section. Here, the direct evaluation will be referred to by MMD-VAE, approximate inference method by MMD-VAE + MCMC, and future prediction by MMD-VAE + FP. 

We also compare our unsupervised methods to frame-level tool presence predictions from $4$ \emph{supervised} methods. 
Twinanda et al. \cite{Twinanda2016EndoNetAD} use a supervised CNN based on AlexNet~\cite{NIPS2012_4824} to perform tool presence detection in a multi-task manner. 
Sahu et al. \cite{Sahu2016ToolAP} use a transfer learning approach to combine ImageNet~\cite{imagenet} features with time-series analysis to detect tool presence.
Raju et al.\cite{Raju2016ToolAP} combine features from GoogleNet~\cite{googlenet} and VGGNet~\cite{vggnet} for tool presence detection. 
Jin et al.\cite{jin2018tool} use a supervised region-based CNN to spatially localize tools and use these detections to drive frame-level tool presence detection. Results are summarized in Table~\ref{tab:summary}.

\begin{figure}[t]
\centering
\subfloat[MMD-VAE\label{fig:mmd-tsne}]{%
\includegraphics[width=0.45\textwidth]{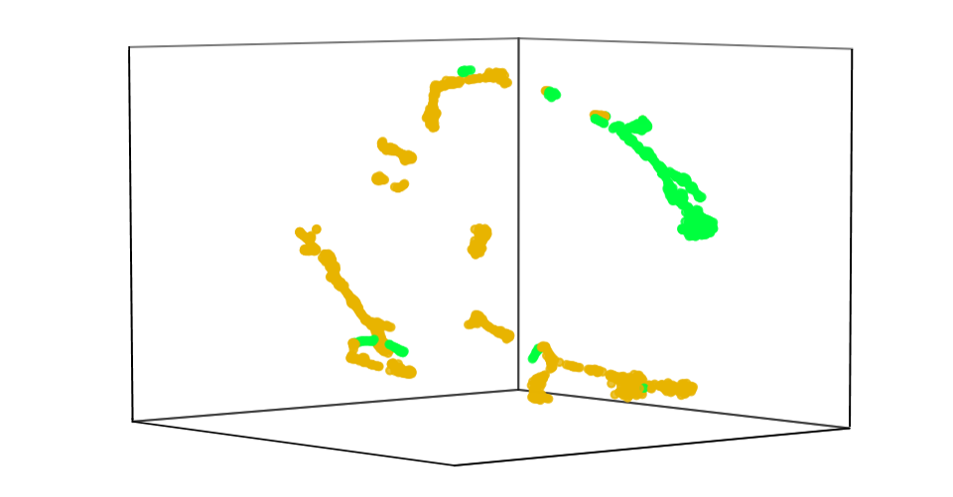}%
}\hfil
\subfloat[MMD-VAE + FP\label{fig:mmdfp-tsne}]{%
\includegraphics[width=0.45\textwidth]{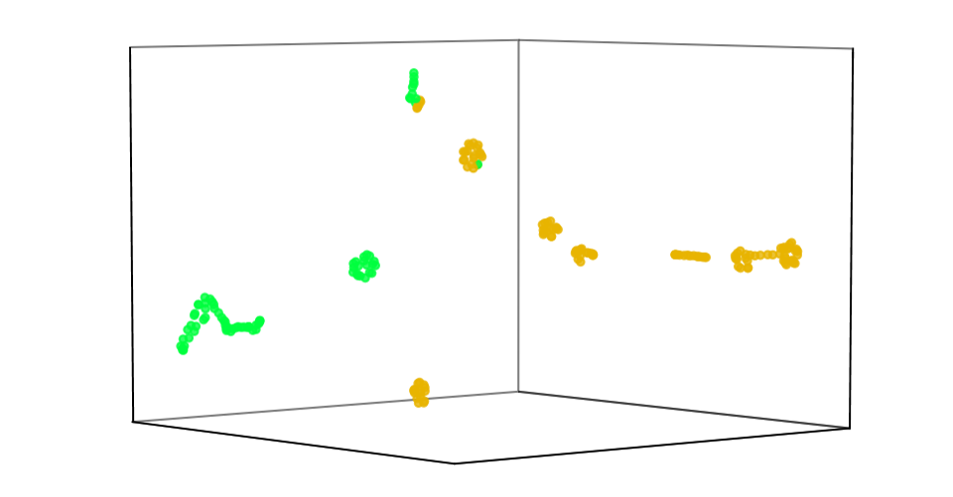}%
}
\caption{3D dimensionality reductions, obtained using t-SNE, of (a) 80D encodings produced by MMD-VAE, and (b) 64D encodings produced by MMD-VAE + FP. MMD-VAE + FP shows slightly better separation between tool (green) and no tool (orange). The labels are used for visualization only.}
\label{fig:tsne}
\end{figure}

\begin{table}[b]
\centering
\caption{Average Precision (AP) in frame-level detection of tool presence \newline(* indicates supervised method)}\label{tab:summary}
\begin{tabular}{l@{\hskip 0.5cm}c}
\hline
   Twinanda et al.* & 52.5\\
   Sahu et al.* & 54.5\\
   Raju et al.*  & 63.7\\
   Jin et al.* & 81.8\\
   MMD-VAE & 71.56\\
   MMD-VAE + MCMC & 73.93\\
   MMD-VAE + FP & 76.18\\
   \hline
\end{tabular}
\end{table}

\subsubsection{Direct Evaluation.} 
The encodings produced by our implementation of MMD--VAE show some amount of separation (Fig.~\ref{fig:tsne}a).
Therefore, we expect our queries to produce high response when evaluated against frames with tools. 
However, we also expect queries to produce higher response against nearby frames with tools compared to frames that are further away from the queries. This is because our per frame encodings may not be able to disentangle the presence and absence of tools over the varied anatomy present in our video sequences (Fig.~\ref{fig:examples}). Direct evaluation achieves an AP of $71.56$ in identifying frames with tools.

\subsubsection{Approximate Inference.} 
\begin{figure}[b]
    \centering
    \includegraphics[width=0.95\textwidth]{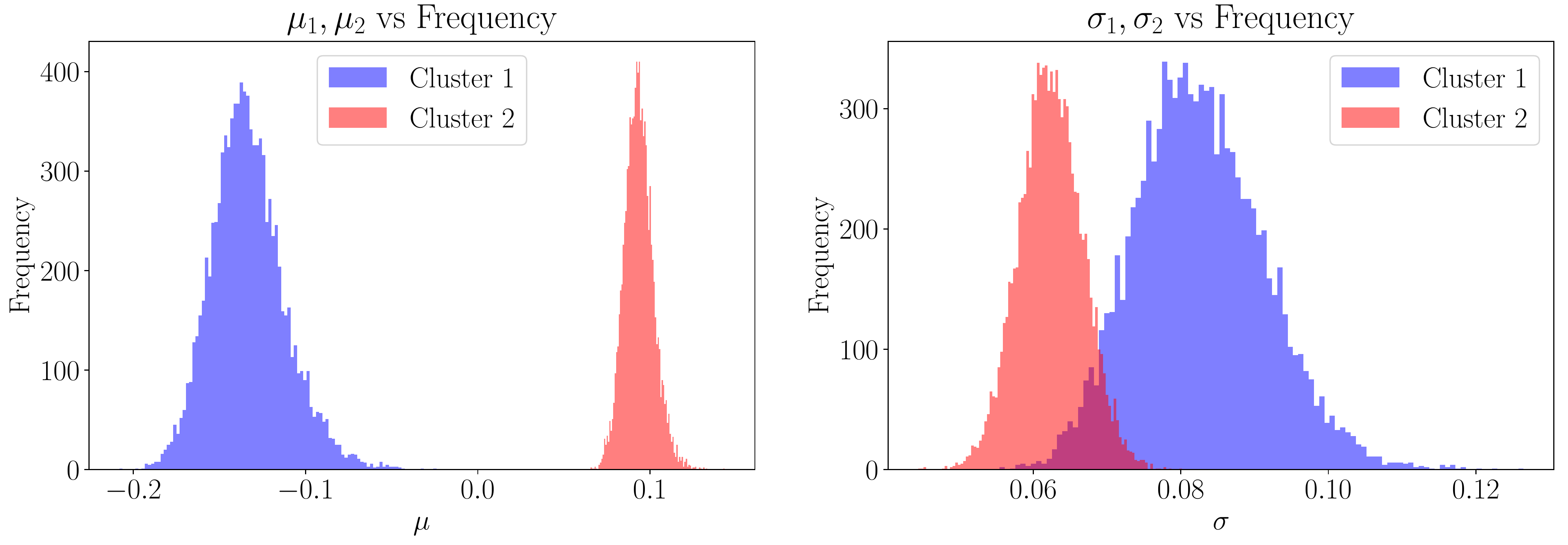}
    \caption{Latent cluster parameter summaries for trained MMD-VAE + MCMC model with means (left) and standard deviations (right). The two configurations of $C$ are described by $\mathcal{N}(-0.13, 0.0068)$ and $\mathcal{N}(0.093, 0.0039)$.}
    \label{fig:mm_separation}
\end{figure}

Since the encodings from the MMD--VAE show some separation,
we expect a trained mixture of two Gaussians to capture the separation in the encodings
and generalize to labeling our held-out test set. We found that the trained mixture model learned two clusters with distinct means and no overlap (Fig.~\ref{fig:mm_separation}). 
By treating each cluster as a binary tool indicator, we can predict labels for the encodings in test set. Compared to the direct evaluation method, we show improvement with an AP of $73.93$.  We hypothesize that our two assumptions that (1) the data can be represented as a mixture of two Gaussians, and (2) a sample belonging to a hidden configuration directly indicates tool presence or absence may be too strong and, therefore, limiting the improvement in AP.
Instead, the clusters likely capture a combination of tool presence and anatomy variance, and precision could be further improved by either relaxing the assumptions or increasing the model complexity.

\subsubsection{Future Prediction.}We expect the addition of temporal information to allow for better disentanglement between tool motion from camera motion. We observe this improvement in the slightly greater separation in the encodings produced by MMD-VAE + FP than those produced without FP (Fig.~\ref{fig:tsne}b). This translates to further improvement in AP at $76.18$. 
We hypothesize that larger gains in AP were again limited due to the short $5$ frame sequence of encodings used to train the future prediction network. Using longer sequences may allow detection of tools over larger variations in anatomy and, therefore, improve overall results.

\section{Conclusion and future work}

We showed through our evaluations that it is possible to learn representations of endoscopic videos that allow us to identify surgical tool presence without supervision. 
We are able to detect frames containing tools directly from MMD--VAE encodings with an AP of $71.56$. By performing approximate inference on these encodings, we are able to improve the AP of frame-level tool presence detection to $73.93$. Finally, using the MMD--VAE encodings to perform future prediction allows us to further improve our AP to $76.18$. 

These results are comparable to those achieved by prior supervised methods evaluated on the M2cai16-tool dataset\cite{Twinanda2016EndoNetAD}. This dataset consists of 15 videos of cholecystectomy procedure, where each frame is labeled with the presence or absence of seven possible surgical tools in a multi-label fashion. As this work was evaluated on a a different dataset, our immediate next step is to re-evaluate our unsupervised method on the M2cai16-tool dataset. This comparison will not only allow us to better understand how our methods compare against supervised methods, but also allow us to evaluate whether our methods can learn generalized representations across various surgical tools.

Going forward, we will explore whether variations in latent dimension and regularization for training our MMD-VAE can improve our ability to discriminate between frames with and without tools. 
We will also explore whether classification can be improved by accommodating the variance in anatomy by relaxing the assumption that our encodings are a mixture of two Gaussian states. Another space to explore will be whether larger sequences of video frame encodings allow us to better separate tool motion from camera motion. Although our initial work is on a limited endoscopic video dataset, our results are promising and our method can be easily applied to larger datasets with wider range of tools and anatomy since we do not rely on labels for training. 

The ability to reliably identify frames containing tools can help the annotation process and can also enable further research in many different areas. For instance, methods that rely on endoscopic video frames without tools~\cite{Liu18} can easily discard frames that are labeled as containing tools. Further, by treating features like optical flow vectors from sequences of frames with and without tools differently, we can work on identifying pixels containing tools without supervision. Unsupervised segmentation of tools, in turn, can enable unsupervised tool tracking and can have great impact on research toward video-based surgical activity recognition and skill assessment.

\subsubsection*{Acknowledgements}
This work was supported by the Johns Hopkins University Provost's Postdoctoral fellowship, NVIDIA GPU grant, and other Johns Hopkins University internal funds.
We would also like to thank Daniel Malinsky and Robert DiPietro for their invaluable feedback. We would also like to acknowledge the JHU Department of Computer Science providing a research GPU cluster.

\bibliographystyle{splncs04}
\bibliography{mybib}

\begin{thebibliography}{10}
\providecommand{\url}[1]{\texttt{#1}}
\providecommand{\urlprefix}{URL }
\providecommand{\doi}[1]{https://doi.org/#1}

\bibitem{Attia17}
{Attia}, M., {Hossny}, M., {Nahavandi}, S., {Asadi}, H.: {Surgical tool
  segmentation using a hybrid deep CNN-RNN auto encoder-decoder}. In: 2017 IEEE
  International Conference on Systems, Man, and Cybernetics (SMC). pp.
  3373--3378 (Oct 2017)

\bibitem{cho14}
Cho, K., van Merri{\"e}nboer, B., Gulcehre, C., Bahdanau, D., Bougares, F.,
  Schwenk, H., Bengio, Y.: Learning phrase representations using {RNN}
  encoder{--}decoder for statistical machine translation. In: Proc. Conference
  on Empirical Methods in Natural Language Processing ({EMNLP}). pp. 1724--1734
  (2014)

\bibitem{imagenet}
{Deng}, J., {Dong}, W., {Socher}, R., {Li}, L., {Kai Li}, {Li Fei-Fei}:
  Imagenet: A large-scale hierarchical image database. In: 2009 IEEE Conference
  on Computer Vision and Pattern Recognition. pp. 248--255 (June 2009).
  \doi{10.1109/CVPR.2009.5206848}

\bibitem{Dipietro18}
DiPietro, R., Hager, G.D.: {Unsupervised Learning for Surgical Motion by
  Learning to Predict the Future}. In: Medical Image Computing and Computer
  Assisted Intervention. pp. 281--288 (2018)

\bibitem{Dipietro16}
DiPietro, R., Lea, C., Malpani, A., Ahmidi, N., Vedula, S.S., Lee, G.I., Lee,
  M.R., Hager, G.D.: {Recognizing Surgical Activities with Recurrent Neural
  Networks}. In: Medical Image Computing \& Computer-Assisted Intervention. pp.
  551--558 (2016)

\bibitem{ephrat}
Ephrat, M.: Acute sinusitis in {HD} (2013),
  \url{www.youtube.com/watch?v=6niL7Poc\_qQ}

\bibitem{Herrera17}
Garc{\'i}a-Peraza-Herrera, L.C., Li, W., Gruijthuijsen, C., Devreker, A.,
  Attilakos, G., Deprest, J., Vander~Poorten, E., Stoyanov, D., Vercauteren,
  T., Ourselin, S.: {Real-Time Segmentation of Non-rigid Surgical Tools Based
  on Deep Learning and Tracking}. In: Computer-Assisted and Robotic Endoscopy
  (CARE). pp. 84--95 (2017)

\bibitem{Gers00}
Gers, F.A., Schmidhuber, J., Cummins, F.A.: Learning to forget: Continual
  prediction with {LSTM}. Neural Computation  \textbf{12},  2451--2471 (2000)

\bibitem{Hochreiter97}
Hochreiter, S., Schmidhuber, J.: Long short-term memory. Neural Computation
  \textbf{9}(8),  1735--1780 (1997)

\bibitem{hoffman2014no}
Hoffman, M.D., Gelman, A.: The {No-U}-turn sampler: Adaptively setting path
  lengths in {Hamiltonian Monte Carlo}. J Mach Learn Res  \textbf{15}(1),
  1593--1623 (2014)

\bibitem{jin2018tool}
Jin, A., Yeung, S., Jopling, J., Krause, J., Azagury, D., Milstein, A.,
  Fei-Fei, L.: Tool detection and operative skill assessment in surgical videos
  using region-based convolutional neural networks. IEEE Winter Conference on
  Applications of Computer Vision  (2018)

\bibitem{vggnet}
Karen~Simonyan, A.Z.: Very deep convolutional networks for large-scale image
  recognition. ArXiv  \textbf{abs/1409.1556} (2014)

\bibitem{kingma2014adam}
Kingma, D.P., Ba, J.: Adam: A method for stochastic optimization.
  arXiv:1412.6980  (2014)

\bibitem{kingma2013auto}
Kingma, D.P., Welling, M.: {Auto-Encoding Variational Bayes}. {arXiv:1312.6114}
   ({2013})

\bibitem{NIPS2012_4824}
Krizhevsky, A., Sutskever, I., Hinton, G.E.: Imagenet classification with deep
  convolutional neural networks. In: Pereira, F., Burges, C.J.C., Bottou, L.,
  Weinberger, K.Q. (eds.) Advances in Neural Information Processing Systems 25,
  pp. 1097--1105. Curran Associates, Inc. (2012),
  \url{http://papers.nips.cc/paper/4824-imagenet-classification-with-deep-convolutional-neural-networks.pdf}

\bibitem{Lea16}
{Lea}, C., {Vidal}, R., {Hager}, G.D.: {Learning Convolutional Action
  Primitives for Fine-grained Action Recognition}. In: {2016 IEEE International
  Conference on Robotics and Automation (ICRA)}. pp. 1642--1649 (May 2016)

\bibitem{Liu18}
Liu, X., Sinha, A., Unberath, M., Ishii, M., Hager, G.D., Taylor, R.H., Reiter,
  A.: {Self-supervised Learning for Dense Depth Estimation in Monocular
  Endoscopy}. In: Computer Assisted Robotic Endoscopy (CARE). pp. 128--138
  (2018)

\bibitem{Malpani15}
Malpani, A., Vedula, S.S., Chen, C.C.G., Hager, G.D.: A study of crowdsourced
  segment-level surgical skill assessment using pairwise rankings.
  International Journal of Computer Assisted Radiology and Surgery
  \textbf{10}(9),  1435--1447 (Sep 2015)

\bibitem{murphy2012machine}
Murphy, K.P.: Machine learning: a probabilistic perspective. MIT press (2012)

\bibitem{Pakhomov17}
Pakhomov, D., Premachandran, V., Allan, M., Azizian, M., Navab, N.: {Deep
  Residual Learning for Instrument Segmentation in Robotic Surgery}.
  {arXiv:1703.08580}  ({2017})

\bibitem{paszke2017automatic}
Paszke, A., Gross, S., Chintala, S., Chanan, G., Yang, E., DeVito, Z., Lin, Z.,
  Desmaison, A., Antiga, L., Lerer, A.: Automatic differentiation in pytorch.
  In: NIPS-W (2017)

\bibitem{Raju2016ToolAP}
Raju, A., Wang, S., Huang, J.: M2cai surgical tool detection challenge report
  (2016)

\bibitem{Sahu2016ToolAP}
Sahu, M., Mukhopadhyay, A., Szengel, A., Zachow, S.: Tool and phase recognition
  using contextual cnn features. ArXiv  \textbf{abs/1610.08854} (2016)

\bibitem{Shvets18}
{Shvets}, A.A., {Rakhlin}, A., {Kalinin}, A.A., {Iglovikov}, V.I.: {Automatic
  Instrument Segmentation in Robot-Assisted Surgery using Deep Learning}. In:
  17th IEEE Int Conference on Machine Learning and Applications (ICMLA). pp.
  624--628 (2018)

\bibitem{Srivastava15}
Srivastava, N., Mansimov, E., Salakhutdinov, R.: Unsupervised learning of video
  representations using lstms. In: Proc. 32nd Int Conference on International
  Conference on Machine Learning. ICML'15, vol.~37, pp. 843--852. JMLR.org
  (2015)

\bibitem{pystan}
{Stan Development Team}: {PyStan: the Python interface to Stan, Version
  2.17.1.0.} (2018), \url{http://mc-stan.org}

\bibitem{googlenet}
Szegedy, C., Liu, W., Jia, Y., Sermanet, P., Reed, S., Anguelov, D., Erhan, D.,
  Vanhoucke, V., Rabinovich, A.: Going deeper with convolutions. In: Computer
  Vision and Pattern Recognition (CVPR) (2015),
  \url{http://arxiv.org/abs/1409.4842}

\bibitem{Tsui13}
Tsui, C., Klein, R., Garabrant, M.: Minimally invasive surgery: national trends
  in adoption and future directions for hospital strategy. Surgical Endoscopy
  \textbf{27}(7),  2253--2257 (Jul 2013)

\bibitem{Twinanda2016EndoNetAD}
Twinanda, A.P., Shehata, S., Mutter, D., Marescaux, J., de~Mathelin, M., Padoy,
  N.: Endonet: A deep architecture for recognition tasks on laparoscopic
  videos. IEEE Transactions on Medical Imaging  \textbf{36},  86--97 (2016)

\bibitem{Zhao2017InfoVAEIM}
Zhao, S., Song, J., Ermon, S.: {InfoVAE: Information Maximizing Variational
  Autoencoders}. {arXiv:1706.02262}  ({2017})

\bibitem{zhu2004recall}
Zhu, M.: Recall, precision and average precision. Department of Statistics and
  Actuarial Science, University of Waterloo, Waterloo  \textbf{2}, ~30 (2004)

\end{thebibliography}
\end{document}